\documentclass{article}

\usepackage[final]{neurips_2022}

\usepackage[utf8]{inputenc} 
\usepackage[T1]{fontenc}    
\usepackage{hyperref}       
\usepackage{url}            
\usepackage{booktabs}       
\usepackage{amsfonts}       
\usepackage{nicefrac}       
\usepackage{microtype}      
\usepackage{xcolor}         
\usepackage{natbib}
\usepackage{graphicx}
\usepackage{amsmath}

\title{Design Process is a Reinforcement Learning Problem}

\author{%
  Reza kakooee* \\
  Department of Architecture\\
  ETH Zurich\\
  Switzerland \\
  \texttt{kakooeer@ethz.ch} \\
  \And
  Benjamin Dillunberger \\
  Department of Architecture\\
  ETH Zurich\\
  Switzerland \\
  \texttt{dillenburger@arch.ethz.ch} \\
}

\begin{document}

\maketitle

\begin{abstract}
  While reinforcement learning has been used widely in research during the past few years, it found fewer real-world applications than supervised learning due to some weaknesses that the RL algorithms suffer from, such as performance degradation in transitioning from the simulator to the real world. Here, we argue the design process is a reinforcement learning problem and can potentially be a proper application for RL algorithms as it is an offline process and conventionally is done in CAD software - a sort of simulator. This creates opportunities for using RL methods and, at the same time, raises challenges. While the design processes are so diverse, here we focus on the space layout planning (SLP), frame it as an RL problem under the Markov Decision Process, and use PPO to address the layout design problem. To do so, we developed an environment named RLDesigner, to simulate the SLP. The RLDesigner is an OpenAI Gym compatible environment that can be easily customized to define a diverse range of design scenarios. We publicly share the environment to encourage both RL and architecture communities to use it for testing different RL algorithms or in their design practice. The codes are available in the following GitHub repository \href{https://github.com/RezaKakooee/rldesigner/tree/Second_Paper} {\url{https://github.com/RezaKakooee/rldesigner/tree/Second_Paper}}.
\end{abstract}

\section{Introduction} \label{I}
 

Reinforcement learning (RL) has been found as a proper framework to support a diverse range of applications over the past few years such as playing video games \citep{shao2019survey}, controlling dynamic systems \citep{recht2019tour}, finance \citep{liu2021finrl}, healthcare \citep{yu2021reinforcement}, and self-driving cars \citep{kiran2021deep}. Although these applications are so diverse, what they share in common is that they are sequential decision processes (SDP). This paper argues that the design process is a reinforcement learning problem because the design process is an SDP as well. Nonetheless, applying RL methods in the design process has grasped less attention so far. This paper explains the similarities between the design process and the RL problem to open up new opportunities for RL researchers and practitioners to use RL algorithms to support different design processes.
 

Design is a key step in making physical things, yet it is difficult to find a single definition for the design process. This paper follows Christopher Alexander’s definition of the design process as a guideline because his rich theories have been used in various domains from architectural design to software \citep{alexander1964notes}. Alexander believes that every design process is involved finding a fitness between two intangible components: the form and its context. The context defines the design problem and the form offers a solution for that problem, meaning that the designer’s goal is to create a form that fits various requirements of its context. The design process includes an initial form that is born in the context, and altering the form in a series of steps until reaching the desired design solution. 

The reinforcement learning problem is procedurally similar to the design process. RL contains an agent that interacts with its environment to learn how to create a behavior that maximizes its long-term reward. The analogy between the RL problem and the design process can be touchable if we consider the RL agent as an artificial designer, the environment as the context, and the learned behavior as the form. Environment and context are entities that determine the boundaries of the learning and design process, respectively. Both RL and the design process are concerned with optimizing a notion of the objective function. 

The real challenge in the design process is that we want to find harmony between two intangible components: a form that has not been designed yet, and a context that cannot be fully described. Similarly, in RL we need an optimal policy to maximize the reward but we have not yet learned the policy and often we do not access the environment’s model that generates the reward. 

Despite these similarities, the application of RL methods in the design process has not yet been well-explored. Since the design processes are so diverse, this paper focuses on a specific design task called Space Layout Planning (SLP) which can be considered as a constrained combinatorial optimization problem \citep{michalek2002architectural}, and attempts to frame it as an RL problem under the Markov Decision Process (MDP). We introduce a new approach for layout design and describe different components of the MDP including state-space, action-space, and reward functions. We further use Proximal Policy Optimization (PPO) \citep{schulman2017proximal} which is one of the state-of-the-art deep RL algorithms to address space layout design. 


Spatial layout planning concerns shaping, arranging, and dimensioning spacial elements to satisfy geometrical, topological, and performance constraints while following certain objectives \citep{jo1995genetic}. It finds patterns of 2D subdivisions on urban, building, apartment (floor plan), and even on element (room, wall, etc.) scales. Similar geometric challenges appear in all these scales for subdividing an existing region into parts to achieve a specific set of properties, such as predefined areas, neighborhoods, proportions (aspect ratio of rooms), etc. This paper aims to address this kind of problem by strategically focusing on the floor plan scale.

In the floor plan scale, the SLP is an important part of the early design phase because it highly affects the building efficiency and end-user comfortability and determines the top-level spatial structure of a building from the early stages of the design process \citep{granadeiro2013building}. Due to innumerable alternatives in configuration, finding feasible layout design solutions falls in the NP-complete category, for which finding an optimal solution might be impossible \citep{gero1998evolving}.


Framing the design process as an RL problem could provide opportunities for RL methods as the design process is an offline process so we do not need to be worried about the real-time reactions of the RL agent. Also, designers conventionally use CAD software which can be considered as an environment simulator. Thus, the RL algorithm we use will not suffer from simulation to real-world performance degradation as there is no need to run the agent outside of the simulator.
 
The design processes have some characteristics that challenge RL methods. For instance, for a specific design problem, there could be endless design scenarios. This is challenging because RL methods usually suffer from generalization abilities that make it difficult to transfer knowledge between different design scenarios. Moreover, for any design scenario, there could be multiple solutions that might prevent the training process from efficiently transitioning from exploration to exploitation as the agent could switch between different solutions during the training process. Nonetheless, this paper aims to put RL methods into the design practice to study its performance hoping to open up new research directions for both RL and architecture communities.

\section{Related works} \label{II}

Several approaches have been proposed in the literature to support the automatic generation of floor plan layouts, including but not limited to shape grammars \citep{stiny1980introduction}, graph synthesis \citep{levin1964use}, and evolutionary strategies \citep{dillenburger2009building}. A good overview of those can be found in these two papers \citep{calixto2015literature} and \citep{veloso2019multi}.

Moreover, numerous attempts have emerged over the last few years to apply machine learning algorithms to architectural applications such as supervised Bayesian networks to design residential building layouts \citep{merrell2010computer}, self-organizing maps for clustering floor plans \citep{dillenburger2016raumindex}, interior design \citep{racec2016computational}, measuring similarities between architectural designs \citep{yoshimura2019deep}, 3D architectural form style transfer \citep{zhang20203d}, topology optimization of building facades \citep{bernhard2020topology}, and house style recognition \citep{yi2020house}.

Generative Adversarial Network (GAN) \citep{goodfellow2014generative} and its different extensions have been also utilized to design floor plans, such as generating residential plans in the style of the architect Le Corbusier \citep{newton2019deep}, generating real-time floor plans incorporated with user interaction \citep{chaillou2019archigan} and creating floor plans for input spatial adjacency graphs \citep{nauata2021house}. Although these approaches can synthesize space layouts according to some design constraints, particularly topological constraints, they tend to generate new samples that are similar to the training examples.

A few studies have been carried out on applying RL in architectural design \citep{xie2013artist, ganin2018synthesizing, bapst2019structured, akizuki2020generative}, and in floor plan design. As one of the first attempts, researchers used RL to learn the best sequence of shape grammars to design floor plans \citep{ruiz2013design} and later extended their approach to designing small single-family dwellings that satisfy habitability and energy efficiency requirements \citep{mandow2020architectural}. Their proposed method, however, highly depends on defining shape grammar rules in the early steps of the training process which is not straightforward.

The Monte-Carlo Tree Search approach was used to design floor plans under dense adjacency and non-adjacency constraints resulting in a scalable approach \citep{shi2020addressing}. However, their framework is limited to only rectangular rooms and plans. Recently, multi-agent deep reinforcement learning (MARL) has been employed to synthesize spatial configuration \citep{veloso2021self}. In their framework, each agent represents a specific spatial partition exploring the environment and communicating with each other to achieve certain objectives related to satisfying some geometrical and topological constraints. However, their approach relies on the assumption that the outline is not restricted to a certain shape and size.

\section{The RLDesigner environment} \label{III}
This section introduces RLDesigner which is the environment we developed to simulate SLP.

\subsection{Markov Decision Process}
Reinforcement learning is a sequential decision process in which the learning happens based on the interaction between the RL agent and its environment. Such interaction is formulated under the Markov Decision Process (MDP) which contains the tuple $(S, A, P, R, \gamma)$, where $S$ represents the state of the environment, $A$ refers to the actions available in each state, $P$ represents the state transition probabilities, $R$ refers to the reward function and $\gamma$ is the discount factor which determines the importance of distant reward compared to immediate reward. In many RL problems, the agent does not have access to $P$ and $R$. As a result, the information the agent uses to make decisions is only $S$ and $R$, and likely the previous action(s). 

\begin{figure}
	\begin{center}
		\includegraphics[width=2.5cm]{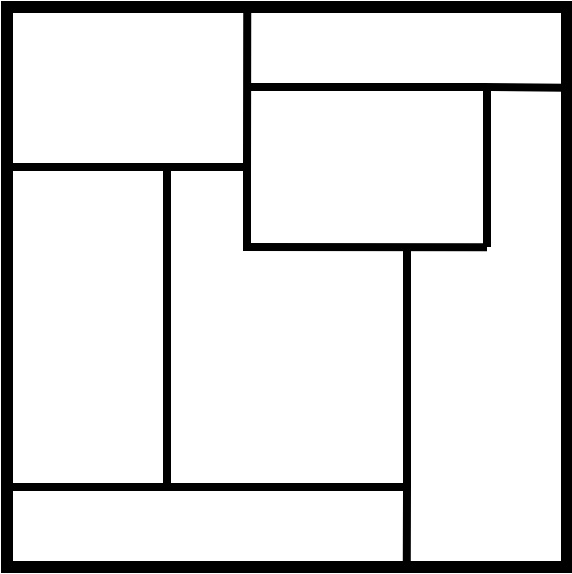}
	\end{center}
	\caption{A sample layout consist of a few space elements}\label{fig:1}
\end{figure}

To define the MDP components for SLP, consider Figure \ref{fig:1} which shows a sample layout. A space layout consists of an outline and some inlines that partition the layout to create some sub-regions. In floor plans, all lines are walls and sub-regions refer to rooms. Such a layout can be designed in two ways: occupying the layout by placing rooms and partitioning the layout by drawing walls \citep{veloso2019multi}. This papers only focus on the partitioning approach as our primary trial-and-error tests showed that it has a better exploratory ability which will be discussed later. 

\begin{figure}
	\begin{center}
		\includegraphics[width=7cm]{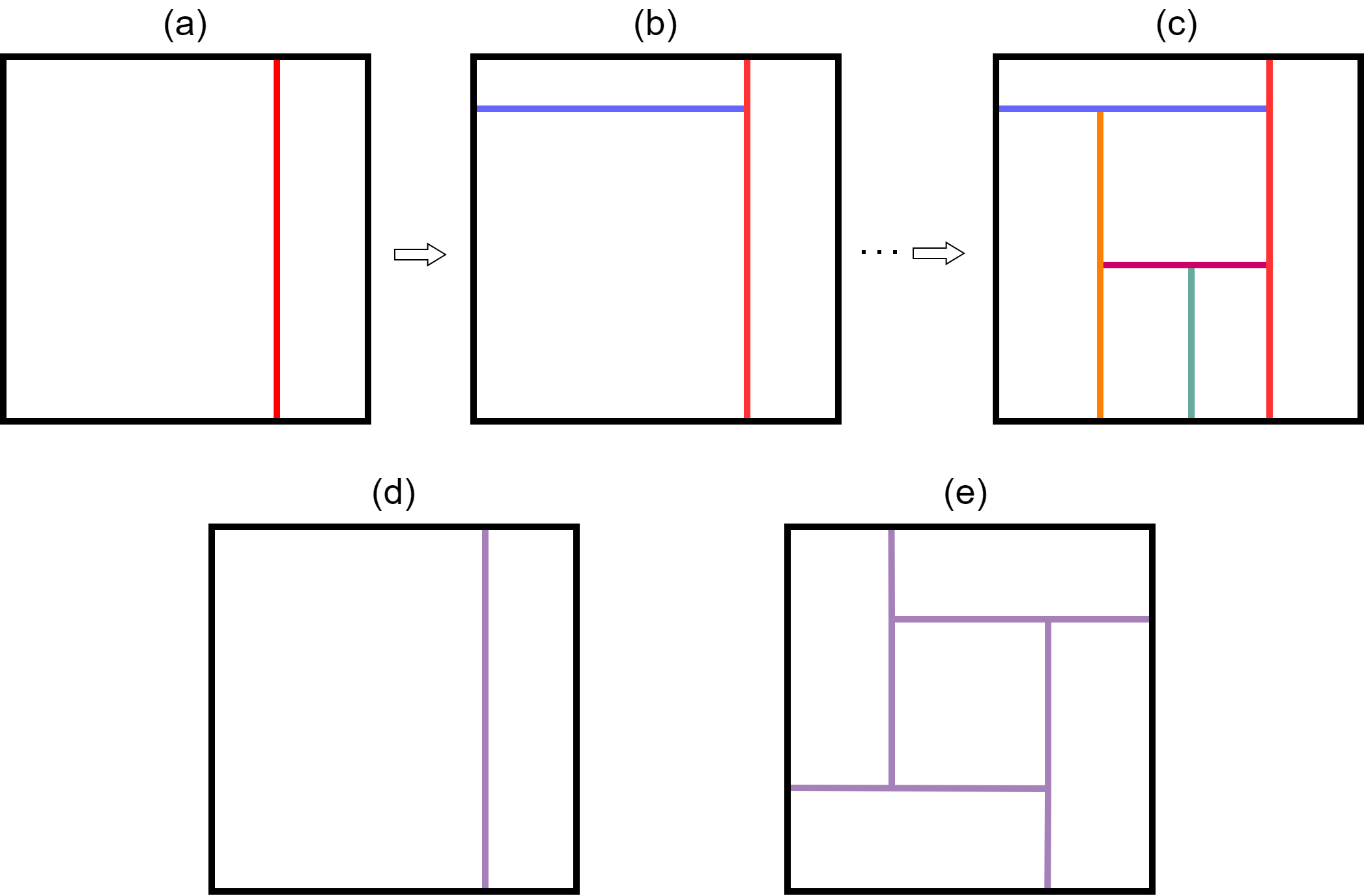}
	\end{center}
	\caption{The first row shows the plan partitioning procedure via rectangular dissection. The second row provides two examples to visualize the shortcomings of rectangular dissection. The left plan shows an unavoidable long wall, and the right one shows a plan that is unachievable by rectangular dissection.}\label{fig:2}
\end{figure}

Traditional partitioning approaches are based on rectangular dissection shown in the first row of Figure \ref{fig:2}. This approach has a few drawbacks: 1- unavoidable long wall starting from one side to the other side of the plan (you will have at least one long wall, see Figure \ref{fig:2}.d), 2- only can create rectangular sub-regions and cannot create concave polygons, 3- it cannot create plans like the one shown in Figure \ref{fig:2}.e \citep{shi2020addressing}.

\begin{figure}
	\begin{center}
		\includegraphics[width=10cm]{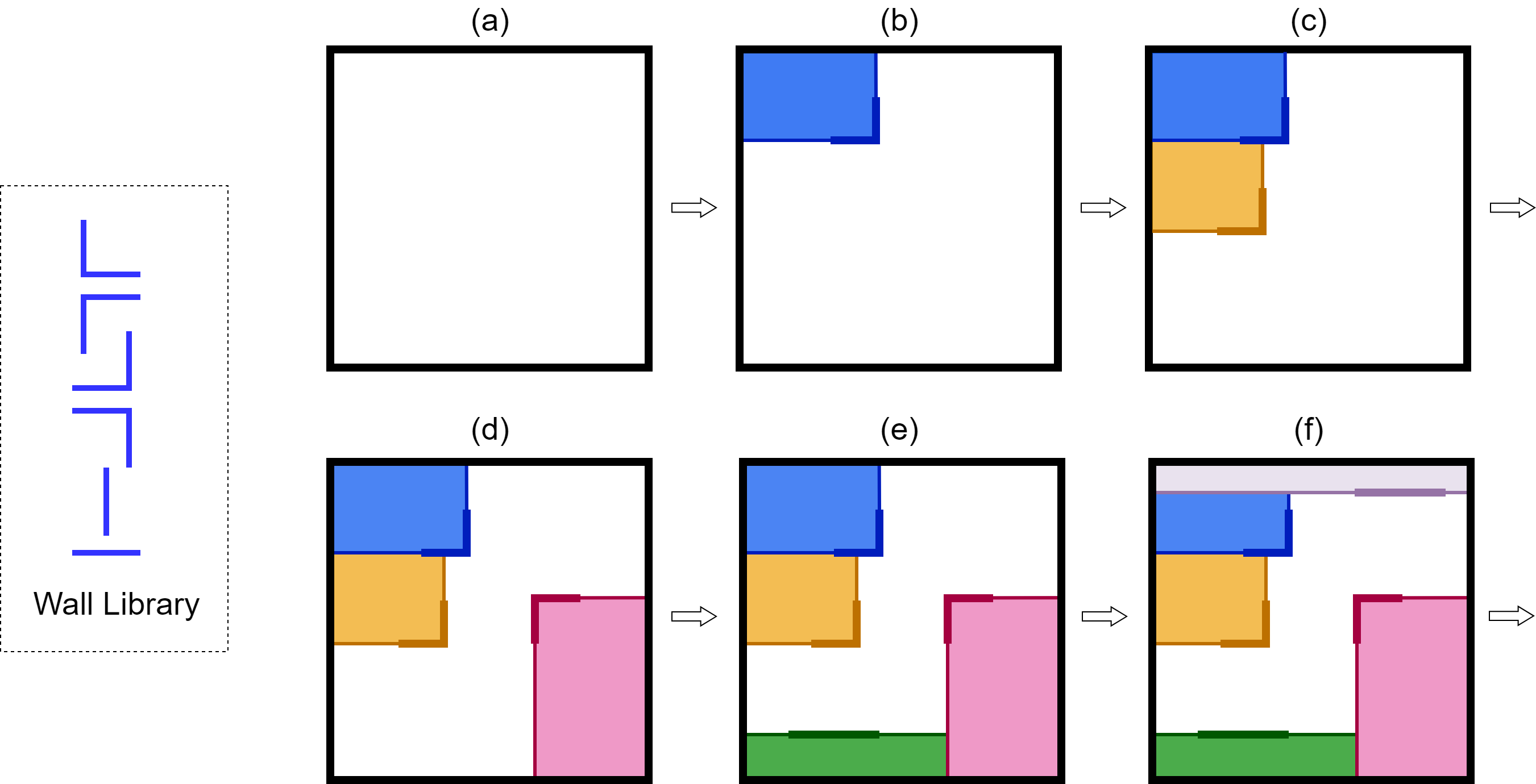}
	\end{center}
	\caption{One-Shot Planning. Partitioning the plan with laser-wall}\label{fig:3}
\end{figure}

In order to overcome these shortcomings, we propose a partitioning approach named laser-wall. Laser-wall is only a fancy name to show some of the properties of our walls. A laser-wall, which can be a straight wall or an angled-wall, consists of two parts: a hard part including two segments making the base of a wall, and a soft part including two segments that extend the base wall to create a complete wall as shown in Figure \ref{fig:3}.b. When a base wall is placed in the plan (the thick segments in Figure \ref{fig:3}.b), it emits the light from both segments to extend them (the thin segments in Figure \ref{fig:3}.b) until the lights hit either a surface with a higher infiltration rate, the hard part of a wall, or the outline. A laser-wall also has a few specifications. First, the soft part of a laser-wall has a certain infiltration rate. The more the infiltration rate of the new wall is the stronger the new light is in cutting the other existing lights. Second, the light cannot infiltrate into the base walls (the hard part of the walls) and the outline. Figure \ref{fig:3}.b shows a base wall and its lights (extensions). By this definition, each laser-wall has five important coordinates: the anchor point which is the center of the base wall, two radiation points which are the front of two segments of the base wall from which the lights radiate, and two endpoints which are the points the lights hit other walls or the outline. By this partitioning approach, we can now explain how to frame this procedure as an MDP. 

The RLDesigner supports both single-agent and multi-agent environments. Here, we only focus on the single-agent setting, and more information about the multi-agent setting can be found in Appendix \ref{AppC}. In the single-agent environment, the design starts from an empty plan and a certain number of required walls. Then, in each timestep, an RL agent places a laser-wall inside the plan to partition the plan. The planning terminates when all walls are being placed. As the single-agent environment includes static walls, we name the single-agent environment as one-shot planning.

\subsection{Action-space}
In the single-agent environment, the action space can be defined as selecting a base wall from the wall library (consisting of six walls as shown in Figure \ref{fig:3}), choosing an infiltration rate (an integer number normalized in $[0, 1, . . . , 9]$), and picking a coordinate in order to place the base wall. After placing the base wall, the light will be emitted automatically from the two segments to make a complete wall. 
Each learning episode terminates when all required walls are placed in the plan. The number of walls depends on the number of rooms we want to create. Figure \ref{fig:3} shows the one-shot planning. Note that the last wall (the top wall in Figure \ref{fig:3}.f) cut the first wall (the blue one) because here we assumed that the infiltration rate of the last wall is greater than the first one.

The number of coordinates depends on the plan resolution which has a high impact on the whole framework performance, as the higher resolution, the plan has, the more accurate design solution will be achieved. Nonetheless, high resolution increases the number of actions as well which can have negative effects on the learning process. For instance, in the single-agent setting one action is to select a coordinate (or cell) to place the selected base wall. Naturally, if the resolution is high, it means that there are more coordinates to choose from.

\subsection{State-space}
The state-space is the playground of the RL agents. The agents live in the state-space and take actions to change the state sequentially, hoping to maximize the cumulative reward. The state-space can be viewed from two perspectives: from the RL agents’ perspective, who take actions according to the state to achieve the design objectives, and from the environment’s perspective to evaluate the agent’s performance based on the quality of the design. Thus, both agents and the environment need to perceive the state to make a decision. 
In the RLDesigner environment, we divide the plan into a 2D grid of discrete cells out of which the state can be defined in two ways: image representation where the state is the RGB image of the plan, and feature vector representation which includes the coordinates of each wall, the distance between five important points of each wall and the other walls, the distance between five important points of each wall and the four main corners of the outline. 

\subsection{Reward function}
In order to define the reward function, we need to understand what a successful design (a good fit) is. The objective of a design is not just the form itself, but the ensemble of form and context \citep{alexander1964notes}. Alexander believes that it is more straightforward to quantify a good fit indirectly by measuring the misfit which explains how much the ultimate design deviates from the desired form. To understand it, assume that the only objective of the design is to reach out to desired areas which can be shown by $A^*$. The misfit can be quantified by the difference between the $A^*$ and the area that the agent just created: $A - A^*$ (where $A$ refers to the area the agent just created). Thus, the reward function can be defined as $-|A - A^*|$, and the goal of the agent would be to maximize the reward which automatically minimizes the misfit. 

Spatial layout design can incorporate several geometrical and topological objectives such as desired area and proportion of rooms, adjacency constraints, day-light requirements, location, orientation of different space components, reachability of each room from the entrance, and so on. In the current version of the RLDesigner environment, we have only incorporated area, proportion, and adjacency. Further criteria will be added in the next versions. According to these criteria, the objective function of the design can be defined as follows:

\begin{equation}
	\begin{aligned}
		\min 		  \quad &  |{Adj} - {Adj}^*|  \\
		\textrm{s.t.} \quad &  A \geq A_{min}     \\
		& |A - A^*|\leq{A_{th}}    \\
		& 1\leq P \leq P^*    \\
	\end{aligned}
\end{equation}

Where $Adj$ and ${Adj}^*$ refer to achieved and desired adjacencies. $A$, $A^*$ and $A_{min}$ refer to the created, desired, and the minimum acceptable areas. $A_{th}$ is the area threshold representing the max difference between the desired and created areas that the environment non-negatively rewards. $P$ and $P^*$ refer to the achieved and desired aspect ratios of rooms.

The way we incorporate this objective function in the training pipeline is as follows. Consider Figure \ref{fig:3}.b as the current state of the environment. Based on this state, the agent takes an action and creates the room shown in Figure \ref{fig:3}.c. Now, let’s assume the difference between the area the agent just created and the desired area is larger than the $A_{th}$. It means that the created area does not satisfy the area constraints. What we do is that we ignore the agent’s action, give a negative reward to the agent, turn the state to Figure \ref{fig:3}.b again, and let the agent to take another action. We follow a similar strategy when the aspect ratio of the created room does not meet the proportion constraints. After some time steps, the agent might create a layout that satisfies all area and proportion constraints. Then, we terminate the episode and give the agent the following reward which is based on the adjacency objective. 
\begin{equation}
	r_{global} = R - M \ge 0
\end{equation}

Where $R$ is a positive, big-enough constant number, and $M$ refers to the number of desired adjacencies that the agent missed while designing the layout. Thus, our setup consists of two types of reward, a local reward which is a binary reward (-1 when the created room does not satisfy the area and/or the proportion constraints, and 0 if it satisfies the constraints), and a global reward which evaluates the ultimate design according to the adjacency objective. We have defined more reward functions which can be found in the repository.

\subsection{Exploratory capabilities of laser-walls}
One of the key challenges when using the RL method is what is called exploration. Exploration conventionally refers to the learning policy’s ability to explore different parts of the environment hoping to find a better sequence of actions that results in more rewards compared to the situation where the agent tends to take only those actions that have been historically proven to be good actions. This form of exploration is related to the behavioral policy of the agent. Nonetheless, exploration could have a secondary form that is related to the environment’s capacity in allowing the RL agent to explore it. Imagine a standard grid-world environment, where the start point is somewhere close to the center of the grid and the goal point is on the top-right. If the agent action set does not include action \textit{right}, no RL algorithm will be able to find the goal point even with a very high exploration rate. This phenomenon happens when the solution space is wide but the action set is limited. 

Spatial layout planning is among those environments that have a large solution space, particularly when the number of spatial components increases. This becomes even worse when the layout itself has a non-convex shape and there are multiple objectives that enlarge the solution space. Thus, finding the optimal solution not only needs more exploration but also requires smart actions that enable the agent to get to the heart of the solution space in order to find the optimal one. The laser-wall concept introduced above is a smart tool in the hand of the RL designer to find a solution in a powerful way. 

As described, the proposed laser-wall has some properties that enable conveniently translating the SLP to an MDP allowing us to deal with this form of planning by RL-based approaches. Laser-wall properties empower the RL agent to partition the plan into diverse architecturally acceptable configurations. It overcomes the challenges that the rectangular dissection approach is faced with, and has the ability to explore a large solution space. The hard segments of the wall ensure the stability of the wall when it is under the exposure of the other walls (as explained the light cannot permeate the hard part of the wall), and the soft part allows the other walls to change the current state of the layout configuration which ease exploring the solution space.

\section{Experiment} \label{IV}
The developed environment is customizable and flexible in generating different design scenarios. The parameters like the number of desired walls (or rooms), desired area for each room, desired aspect ratios (proportions), and desired rooms adjacency matrix can be adapted to generate new design scenarios. The outline of the plan is not always rectangular but can be easily changed to non-rectangular shapes by simply masking one or multiple parts of the plan, as shown in Figure \ref{fig:4}. Masked regions are those the agent is not allowed to place any spatial component in. The size of the plan is also a changeable parameter of the environment. 
In this section, we define a few design scenarios to check both environment flexibility and RL algorithms' performance in reaching the design goal (desired room adjacency matrix) with respect to the design constraints (desired areas and proportions of rooms). 

\begin{table}
	\caption{\label{tab:1}Six random design scenarios for testing the RLDesigner environment.}
	\centering
	\resizebox{\textwidth}{!}{\begin{tabular}{ccccc}
			\toprule
			\textbf{Scenario} & \textbf{\#Rooms} & \textbf{Proportion status} & \textbf{Desired areas ${(m^2)}$} & \textbf{Desired adjacencies}   \\ 
			\midrule
			1   & 4   &  1   & [110, 92, 57, 52]                       &  [[1, 2], [2, 3], [2, 4], [1, 3]]		                    \\ 
			2   & 5   &  1   & [83, 78, 60, 44, 33]                    &  [[1, 2], [1, 5], [3, 4]]		                            \\ 
			3   & 6   &  0   & [111, 75, 38, 34, 30, 21]               &  [[3, 4], [1, 6], [1, 4]]		                            \\ 
			4   & 7   &  1   & [65, 60, 48, 36, 30, 28, 23]            &  [[2, 3], [1, 5], [6, 7], [3, 4]]	                 	    \\ 
			5   & 8   &  0   & [85, 64, 64, 50, 41, 32, 32, 28]        &  [[2, 8], [4, 6], [1, 7], [2, 7]]		                    \\ 
			6   & 9   &  0   & [63, 60, 58, 50, 34, 33, 27, 27, 26]    &  [[3, 9], [8, 9], [3, 6], [7, 8], [5, 7], [1, 8]]		    \\ 
			\bottomrule
	\end{tabular}}
\end{table}

In the scenarios defined in this paper, the plan is always a grid of $20*20$ cells, and can be masked. In half of the scenarios, we did not include proportion as the constraint (in which we set $R = 200$). In the rest, the desired proportion is between $1$ and $P^* = 5$ (in which we set $R = 400$). The last column in the table shows the desired adjacencies as the objective of the design process. For example, $[1, 2]$ shows that there must be a connection between \textit{Room 1} and \textit{Room 2}. The adjacency matrix of the final design might also include some other direct connections which are not a part of the desired adjacencies, but we do not penalize the agent for them. We only penalize the agent when it misses desired adjacencies in the ultimate design. Besides, in all scenarios $A_{min} = 10$, and $A_{th} = 4$. 

\subsection{RL method}
Deep RL consists of various algorithms that can be categorized as model-based RL (MBRL) and model-free RL (MFRL) algorithms according to the availability of the environment's model for the agent. The former represents the prospective goal-directed mechanism in the brain, and the latter represents retrospective habitual learning. The goal-directed mechanism either has access to the environment model or can learn a cognitive map via the interaction with the environment and utilize it for planning \citep{kakooee2021developing}. As in this paper, we encounter a planning problem, and the model of the environment can be available for the agent, both MBRL and MFRL can be used. Nonetheless, we use an MFRL algorithm called Proximal Policy Optimization (PPO) \citep{schulman2017proximal}. 

PPO is a policy gradient method that aims to update the policy parameters gradually by the data collected through interaction with the environment, in contrast to value-based methods that learn a kind of value function to obtain the policy. PPO uses a clipped surrogate objective while ensuring exploration by embedding an entropy term into the objective function. In this paper, we use \textit{Ray-RLlib} \citep{liang2018rllib} implementation of PPO, which supports single-agent and discrete action space settings. The PPO's hyperparameters are the default values of the PPO implementation of the \textit{RLlib}; however, we have adapted the policy network. The network architecture can be found in Appendix \ref{AppA}.

\subsection{Results}
\begin{figure}
	\begin{center}
		\includegraphics[width=15cm]{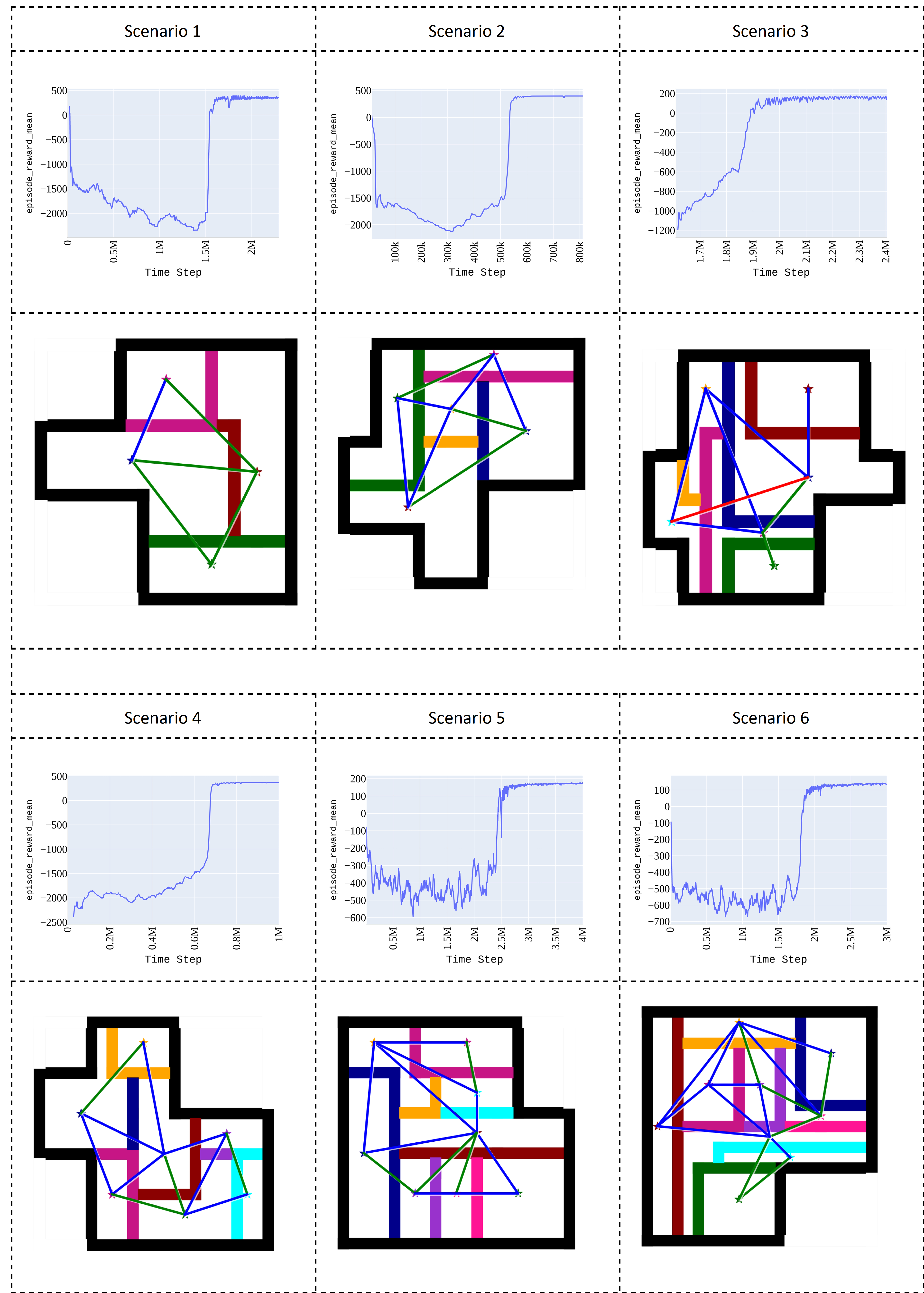}
	\end{center}
	\caption{The learning curves of PPO and the created plans for each scenario defined in Table \ref{tab:1}}\label{fig:4}
\end{figure}

The evaluation of the agent performance is based on the \textit{episode reward mean}, and the number of actions to find the desired solution (e.g. \textit{episode len mean}). Figure \ref{fig:4} shows the performance of the PPO algorithms in the six different scenarios defined in Table \ref{tab:1}. As mentioned above the learning procedure involves maximizing the reward by arranging the room such that the desired adjacency is archived while satisfying the constraints, including the desired room areas and proportions. The first and the third rows of Figure \ref{fig:4} show how learning evolves, and the second and fourth rows display the layout the agent created for each scenario. The thin lines on top of the plans represent the connections between rooms, among which the green lines represent those desired adjacencies which have been successfully achieved, the red lines represent those desired adjacencies that the agent has failed to make, and the blue ones show other connections between rooms. 

As the learning curve shows, the agent has learned a policy that maximizes the reward after enough interaction with the environment. Besides, the episode lengths in all scenarios approach to their minimum values during the training process. The minimum length of each scenario is equal to the number of walls or number of rooms minus one. Since the \textit{episode reward mean} and the \textit{episode len mean} curves are symmetric, we avoid plotting \textit{episode len mean} curves, but as an example see Figure \ref{fig:6} in Appendix \ref{AppB}. Since the longer the episode length is the more negative reward the agent receives, relationship between the two are symmetric. Approaching the episode lengths to their minimum values means that the agent learned to avoid taking those actions that dissatisfy the constraints.

\section{Discussion and future directions} \label{V}
This paper showed that the design process can be considered as an RL problem because like the RL problem, the design process is a sequential decision process as well, and both aim to find a notion of fitness between two intangible components. We studied SLP as a specific design task because SLP is perceived as an important task in various design processes including architectural design practices. Inspired by the existing partitioning approaches, we proposed a novel space partitioning approach and described how an SLP can be framed as an MDP. Consequently, we developed the RLDesigner, a reinforcement learning (RL) environment for general-purpose spatial layout planning. 

We also demonstrated results achieved by training PPO algorithms on some design scenarios to achieve the desired adjacency objectives while satisfying room areas and proportions constraints. The results show how useful RL is in designing the spatial layout. The major goal of the paper was to show that the design process can be formulated as an MDP and RL algorithms are able to design. Nonetheless, there are a lot of investigations that need to be done. First, while we studied the performance of the PPO agent in designing a layout, we did not compare its performance with other RL algorithms and non-RL methods like genetic algorithms, which is a widely used method in SLP, to investigate which one could be a better artificial designer for SLP. Second, for each design scenario mentioned above, we need to train the RL algorithm from scratch. Although the training is relatively fast (see Appendix \ref{AppD}), training from scratch for every new scenario is absolutely inefficient, particularly for complex design scenarios that could consist of many spatial components and multiple constraints and objectives. Nonetheless, this paper is a part of a larger ongoing research project, in which we have the following objectives. 

We already have generated a large dataset of layouts by which we aim to train an offline RL agent to learn a policy for a diverse range of design scenarios. This policy can be either directly used or be fine-tuned for new design scenarios. Since we have access to the environment, fine-tuning the agent could be possible. This strategy might help to enable transfer-learning across design scenarios and equip the learned agent with generalization capabilities. In addition, as the name suggests, SLP is a planning problem and the fact that we have access to the model of the environment allows us to compare model-based and model-free RL algorithms to check which approach can be more efficient to be used as an artificial designer in SLP.

The RLDesigner is an open-source, OpenAI Gym compatible, and customizable environment that allows human designers to adapt different components of the environment to create new design scenarios that satisfy their functional requirements. We are sharing the codes publicly to be used by both RL and architecture communities. For the RL community, it could be a reasonable choice to test both model-based and model-free RL algorithms, also the environment supports both single-agent and multi-agent settings. With simple design scenarios, it can also be useful for teaching purposes as the physics behind the SLP is simple so anyone can understand it easily. Particularly, as the environment is customizable, one can make new design scenarios for various purposes by simply changing some parameters. For the architecture community, we have provided a framework that allows coping with the family of SLP in architecture by a learning-based approach, hoping it could lead to better knowledge sharing, evaluation opportunities, and ultimately to develop more efficient layout design approaches.

\newpage
\vskip 0.2in
\bibliography{neurips_2022}

\section*{Checklist}


\begin{enumerate}

	\item For all authors...
	\begin{enumerate}
		\item Do the main claims made in the abstract and introduction accurately reflect the paper's contributions and scope?
		\answerYes{In Section~\ref{III} we explained how design process can be framed as an RL problem, and in Seciton~\ref{IV} we showed the performance of PPO as an artificial designer.}
		\item Did you describe the limitations of your work?
		\answerYes{See Section~\ref{V}}
		\item Did you discuss any potential negative societal impacts of your work?
		\answerNo{We believe our work does not have any negative societal impacts.}
		\item Have you read the ethics review guidelines and ensured that your paper conforms to them?
		\answerYes{}
	\end{enumerate}

	\item If you are including theoretical results...
	\begin{enumerate}
		\item Did you state the full set of assumptions of all theoretical results?
		\answerNo{We do not have any theoretical results}
		\item Did you include complete proofs of all theoretical results?
		\answerNo{We do not have any theoretical results}
	\end{enumerate}

	\item If you ran experiments...
	\begin{enumerate}
		\item Did you include the code, data, and instructions needed to reproduce the main experimental results (either in the supplemental material or as a URL)?
		\answerYes{We attached the codes as supplemental material but do not share the URL at the moment due to double-blind procedure.}
		\item Did you specify all the training details (e.g., data splits, hyperparameters, how they were chosen)?
		\answerYes{We use the default hyperparameters of the RLlib library. See Section~\ref{IV} and Appendix \ref{AppA}}
		\item Did you report error bars (e.g., with respect to the random seed after running experiments multiple times)?
		\answerNo{We have not performed the experiments for multiple times as the core motivation of the paper is not the algorithmic efficiency but rather is to show how the design process can be a great application for RL methods. Thus, the experimental results only introduce proof-of-concept.}
		\item Did you include the total amount of compute and the type of resources used (e.g., type of GPUs, internal cluster, or cloud provider)?
		\answerYes{See Appendix \ref{AppD}}
	\end{enumerate}

	\item If you are using existing assets (e.g., code, data, models) or curating/releasing new assets...
	\begin{enumerate}
		\item If your work uses existing assets, did you cite the creators?
		\answerYes{We use RLlib library and cited it. See Section~\ref{IV}}
		\item Did you mention the license of the assets?
		\answerYes{The LICENCE can be found in Ray-RLlib GitHub repository.}
		\item Did you include any new assets either in the supplemental material or as a URL?
		\answerYes{We are publicly share the codes of the environemnt we have developed and the RL training pipline which we implemented by using RLlib.}
		\item Did you discuss whether and how consent was obtained from people whose data you're using/curating?
		\answerNo{We do not use any existing datasets.}
		\item Did you discuss whether the data you are using/curating contains personally identifiable information or offensive content?
		\answerNo{We do not use any dataset.}
	\end{enumerate}

	\item If you used crowdsourcing or conducted research with human subjects...
	\begin{enumerate}
		\item Did you include the full text of instructions given to participants and screenshots, if applicable?
		\answerNo{Not applicable.}
		\item Did you describe any potential participant risks, with links to Institutional Review Board (IRB) approvals, if applicable?
		\answerNo{Not applicable.}
		\item Did you include the estimated hourly wage paid to participants and the total amount spent on participant compensation?
		\answerNo{Not applicable.}
	\end{enumerate}

\end{enumerate}


\section*{Appendix}
\appendix

\section{Policy network architecture} \label{AppA}
\begin{figure}
	\begin{center}
		\includegraphics[width=10cm]{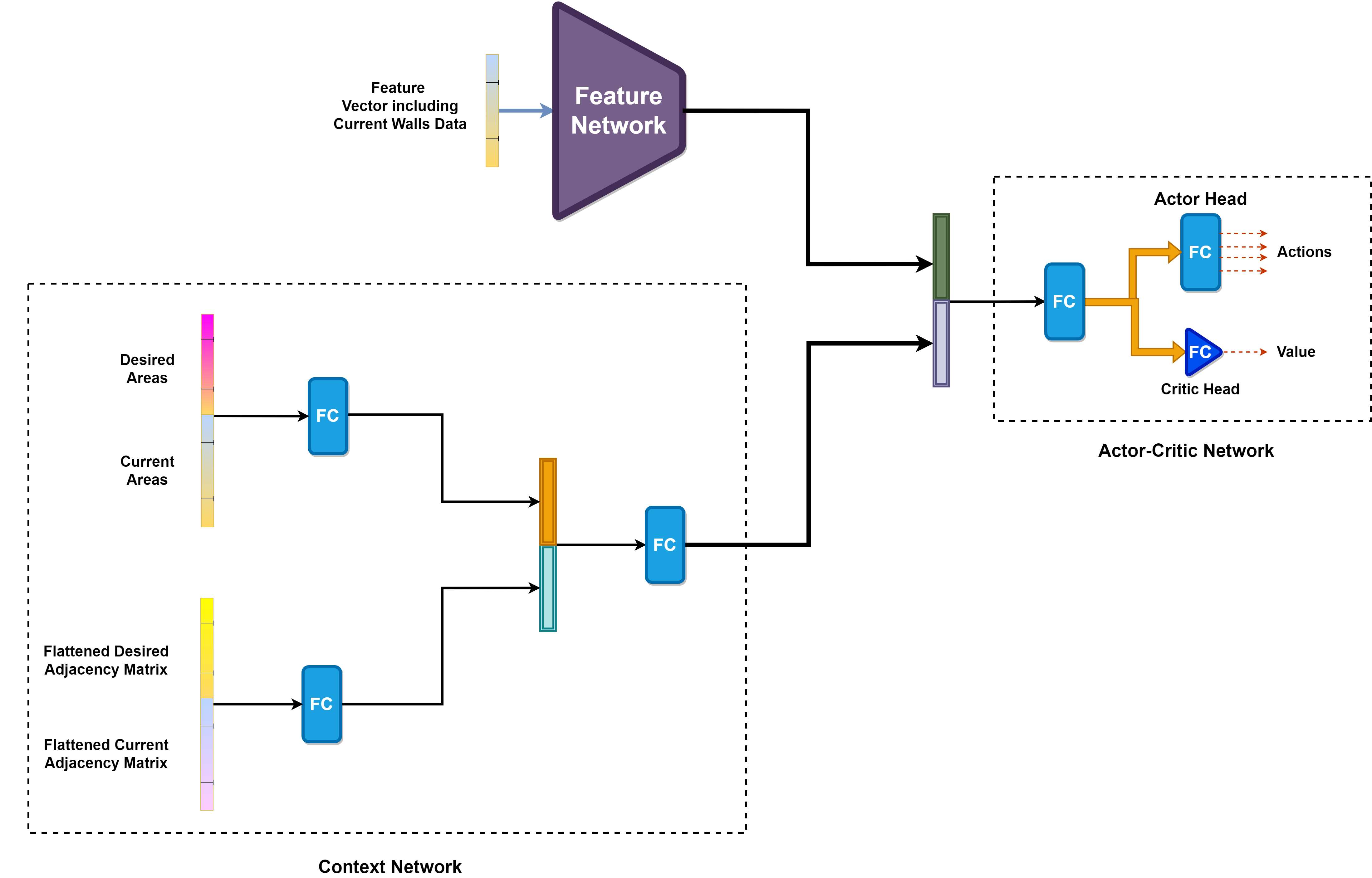}
	\end{center}
	\caption{The policy network consisting of three sub-networks: feature, context, and actor-critic networks}\label{fig:5}
\end{figure}

Our policy network consists of three sub-networks shown in Figure \ref{fig:5}. The first one is a feature network that encodes the current layout into the latent space. It consists of convolutional and linear layers if the state-space is the RGB image of the plan and only linear layers otherwise. 

The second one is a context network whose goal is to encode the design constraints and objectives. As mentioned earlier, there could be endless design scenarios. Thus, we need to inform the agent for which design scenarios we want to find a solution. To do so, we feed the network with both desired areas and adjacencies, and the areas and adjacencies of the current state of the environment. We path them through multiple linear layers and concatenate their outputs with the output of the feature network. We do not incorporate the proportion constraints into the network because proportion constraint is always fixed and does not change across scenarios. Thus, we only check the proportion constraints when evaluating the agent's action. Finally, the third network is the actor-critic model that tries to predict the action based on the embeddings encoded by the first two sub-networks. 

The context network can be ignored if the goal is to train the agent from scratch for each design scenario. In this situation, the agent explores the environment and finds the goal state by chance. Then, the agent tries to exploit the learned trajectory while keeping the exploration alive. A context network will be necessary when the task is to train the agent for multiple design scenarios at once or transfer knowledge across different design scenarios.

\section{Supplementary figures} \label{AppB}
\begin{figure}
	\begin{center}
		\includegraphics[width=6cm]{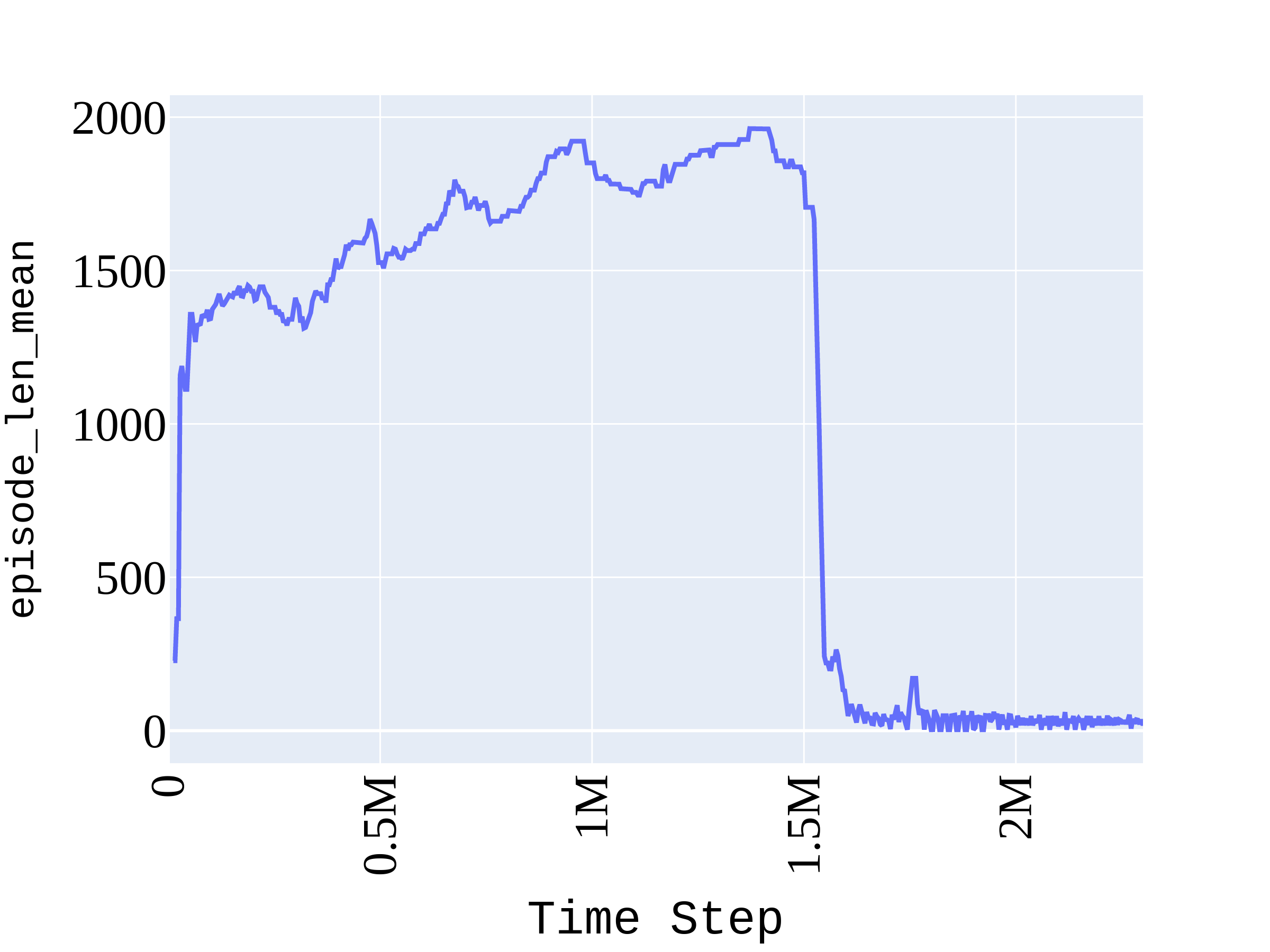}
	\end{center}
	\caption{The \textit{episode len mean} curve for \textit{Scenario 1}.}\label{fig:6}
\end{figure}

Figure \ref{fig:6} shows how the episode length in \textit{Scenario 1} approaches to its minimum value which is $3$ as in this scenario we only have four rooms, so we can design the plan by three laser-walls. 

\section{Multi-agent} \label{AppC}
In the multi-agent environment, each wall represents an agent, and each agent is responsible for creating a room. For example, the kitchen agent is responsible for creating a kitchen with some properties like a $20 m^2$ area. The design starts with an initial plan including some randomly placed laser-walls. Then, in each timestep, an agent takes an action (discussed below) in order to change the plan hoping to make it better with respect to the design constraints and objectives. Thus, the multi-agent environment consists of dynamic laser-walls which can be transformed within the layout. The multi-agent planning terminates when all design constraints are met and the objectives are satisfied.

\subsection{Action-space}
\begin{figure}
	\begin{center}
		\includegraphics[width=8.5cm]{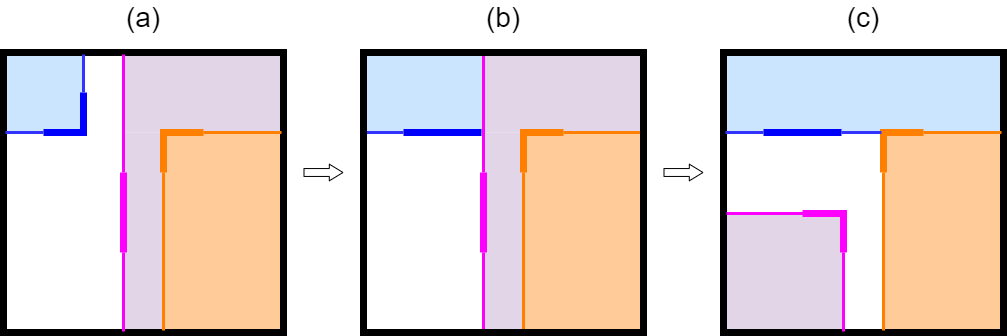}
	\end{center}
	\caption{Multi-Agent Designing. (a): The layout design begins with three walls placed in random coordinates with this order: brown, magenta, and blue. Each agent owns the smaller region that it creates. (b): The blue agent rotated its right segment through ${90^\circ}$ \textit{CW}. (c): The magenta agent rotated its left segment ${90^\circ}$ \textit{CCW}.}\label{fig:7}
\end{figure}

In the multi-agent setting, the planning begins with some laser-walls randomly placed within the plan in a predefined order. The learning process includes transforming the base walls in order to reach the design objectives. Each agent can take one of the following actions in a timestep: rotating the base wall through ${\mp90^\circ}$, rotating only one segment of the base wall through ${\mp90^\circ}$, moving the base wall one cell to left, right, up, and down, flipping the base wall horizontally and vertically, and choosing an infiltration rate. Figure \ref{fig:7} shows how multi-agent planning is executed when the infiltration rate of all walls is the same.

\subsection{State-space}
\begin{figure}
	\begin{center}
		\includegraphics[width=6cm]{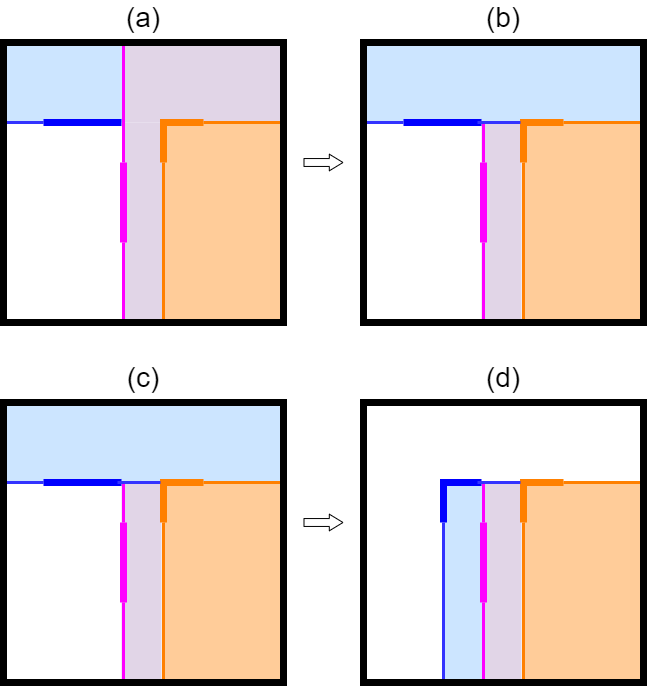}
	\end{center}
	\caption{Challenges of the multi-agent designing. The first row shows how the order of walls can change the ultimate layout. The second row displays the switching effect causing a highly fragile state-space.}\label{fig:8}
\end{figure}

State-space for both single-agent and multi-agent environments are the same; however, in the multi-agent setting the state-space is highly volatile because of two reasons. First, the order the agents take action matters so that the state-space representation could be different even for the same wall configuration as shown in the first row of Figure \ref{fig:8}. The wall configuration is the same in both plans; however, in the left one, the wall’s order is brown, magenta, and blue, while in the right one the order is brown, blue, and magenta. 

Second, each wall partitions the remaining space into two sub-regions and we always assign the smaller area as the corresponding area of that wall. This leads to the situation that we call the switching effect. To understand the switching effect consider the second row of Figure \ref{fig:8} and imagine the left plan is the current state of the environment and it is the blue agent’s turn to take an action. Let’s assume the blue agent decides to rotate its left segment through ${90^\circ}$ \textit{CCW}. Since we assign the small area to the agent, the blue agent owns the blue region shown in the right plan. Thus, the area of the blue agent suddenly switches from the top (Figure \ref{fig:8}.c) to the middle of the plan (Figure \ref{fig:8}.d). 

This high volatility makes the learning process more difficult but creates more opportunities because, firstly, the order of the wall can be embedded in the learning process by defining a master agent (can be also called laser-man) which learns what is the best order for the walls in each state of the environment. Secondly, the switching effect allows the agent to rapidly switch to other parts of the solution space that otherwise might need a lot of actions to be reached. Moreover, the area assigning is a task that can be embedded in the learning process as well so that the agent itself decides to own the small area or the big one. This is a relevant learning task as in many design scenarios we might need to create large rooms.

\subsection{Reward}
The reward in the multi-agent setting is similar to the single-agent. The only difference is that at end of the episode the environment gives the reward to all agents, but within an episode, the reward belongs to only the agent who just took an action, not the rest of the agents.

\section{Computational resources and source codes} \label{AppD}
\label{sec:App}
Training PPO agent for each of the defined scenarios takes about an hour on average on a single GPU machine with NVIDIA GeForce RTX 2080 Ti, AMD® Ryzen threadripper 2920x 12-core processor × 24, and 64 GiB RAM. 

The RLDesigner environment and the PPO agent codes are publicly available on the first author's GitHub.

\end{document}